\documentclass{IOS-Book-Article}

\usepackage{mathtools} 
\usepackage{amssymb}   
\usepackage{txfonts}

\usepackage{soul}\setuldepth{article}
\usepackage{multirow}

\usepackage{times}

\usepackage{hyperref}
\usepackage{url}
\usepackage{xcolor}
\usepackage{amsmath}
\usepackage{graphicx}
\usepackage{url}
\usepackage{tabularx}
\usepackage{dashbox}
\usepackage{color}

\usepackage{subcaption}
\usepackage{comment}

\usepackage{colortbl}

\definecolor{orange}{RGB}{255,170,0}
\definecolor{coralpink}{rgb}{0.97, 0.51, 0.47}
\definecolor{Gray}{gray}{0.85}
\definecolor{LightGray}{gray}{0.91}
\definecolor{yello}{HTML}{D4aa00}
\definecolor{applegreen}{HTML}{0044aa}

\newcommand{\cf}{\emph{cf.}~} 

\newcommand{\norm}[1]{\left\lVert#1\right\rVert}

\def\hb{\hbox to 11.5 cm{}}

\begin{document}
\pagestyle{headings}
\def\thepage{}

\begin{frontmatter}              

\title{Interactively Providing Explanations for Transformer Language Models}



\author[A,B]{Felix Friedrich},
\author[A,B]{Patrick Schramowski},
\author[A]{Christopher Tauchmann}
and
\author[A,B]{Kristian Kersting}

\address[A]{Computer Science Department, TU Darmstadt, Germany}
\address[B]{Hessian Center for AI (hessian.AI)}
{\tt\small \{lastname\}@cs.tu-darmstadt.de}


\begin{abstract}
Transformer language models are state of the art in a multitude of NLP tasks. Despite these successes, their opaqueness remains problematic. Recent methods aiming to provide interpretability and explainability to black-box models primarily focus on post hoc explanations of (sometimes spurious) input-output correlations. Instead, we emphasize using prototype networks directly incorporated into the model architecture and hence explain the reasoning behind the network's decisions. Our architecture performs on par with several language models and, moreover, enables learning from user interactions. This not only offers a better understanding of language models but uses human capabilities to incorporate knowledge outside of the rigid range of purely data-driven approaches.
\end{abstract}

\end{frontmatter}

\section{Introduction}
Transformer language models (LMs) are ubiquitous in NLP today but also notoriously opaque. Therefore, it is not surprising that a growing body of work aims to interpret them: Recent evaluations of saliency methods \cite{ding-koehn-2021-evaluating} and instance attribution methods \cite{pezeshkpour-etal-2021-empirical} find that, while intriguing, different methods assign importance to different inputs for the same outputs. Furthermore, they are usually employed post hoc, thus possibly encouraging reporting bias \cite{10.1145/2509558.2509563}. 
The black-box character of LMs becomes especially problematic, as the data to train them might be unfiltered and contain (human) bias. As a result, ethical concerns about these models arise, which can have a substantial negative impact on society as they get increasingly integrated into our lives \cite{bender2021stochasticparrots__}.

Here, we focus on providing case-based reasoning explanations during the inference process, directly outputting the LM's predictions. In doing so, we address the problems mentioned above of post hoc explanations and help reduce (human) bias. 
To increase the interpretability of the model, we enhance the basic transformer architecture with a prototype layer and propose \textit{Prototypical-Transformer Explanation} (Proto-Trex) Networks\footnote{code available at \url{https://github.com/felifri/XAITransformer}}. Proto-Trex networks provide an explanation as a prototypical example for a specific model prediction, which is similar to (training-)samples with the same label.

To enhance Proto-Trex, we propose an interactive learning setting, iProto-Trex. In addition to simply revealing the network's reasoning by providing prototypical explanations, this approach further enables revising the network's explanations. 
We integrate XIL into prototype networks, which, in contrast to previous (post hoc) XIL methods \cite{XIL_typo}, avoids tracing gradients and only interacts on prototypes (\cf Fig~\ref{fig:3Interaction_pipeline}).
That makes our approach particularly efficient, as users can directly manipulate a network component according to their preferences.
Moreover, this combination is exciting because explanations can be sub-optimal, but the explanation quality is normative, i.e. users have different notions and understandings of a good explanation.


Our experimental results demonstrate that Proto-Trex networks perform on par with non-interpretable baselines, e.g. BERT \cite{devlin2019bert} and GPT \cite{radford2019language}.
More importantly, we show that users can interact with ease by, e.g., simply manipulating the interpretable layer, i.e. a prototype. It enables users of any knowledge to give feedback and manipulate the model regarding their preferences, moving beyond a purely data-driven approach. The improvement through user interaction addresses not only performance in terms of accuracy \cite{ribeiro2016why} but also explanation quality \cite{schramowski2020making} or user trust \cite{teso2019explanatory}.

To summarize, our contributions are as follows: We (i) introduce prototype networks for transformer LMs (Proto-Trex) to provide explanations and (ii) show that they are on par with non-interpretable baselines on classification tasks across different architectures and datasets. Moreover, to improve prototypical explanations, we (iii) provide a novel interactive prototype learning setting (iProto-Trex) accounting for user feedback certainty. 

We proceed as follows. We start by briefly reviewing related work of interpretability in NLP. Then we introduce Proto-Trex networks, including our novel interactive learning setting.
Before concluding, we discuss faithful explanations and touch upon the results of our experimental evaluation.

\section{Towards the Explainability of Transformer Language Models}
To open the black box of transformers, we use i/Proto-Trex motivated by post hoc interpretation methods, case-based reasoning approaches, and explanatory interactive learning.

\paragraph{\bf Post hoc Interpretability.}
Various (post hoc) interpretability methods focus on different parts of the transformer architecture. Several works provide broad overviews of the fast developing field of Explainable AI (XAI) \cite{XAI_overview,atanasova-etal-2020-diagnostic,belinkov-glass-2019-analysis}. Generally, there are methods that analyze word representations \cite{voita-etal-2019-bottom}, the attention distribution throughout the model \cite{jain-wallace-2019-attention,wiegreffe2019attention}, and the (attention and classification) model heads \cite{voita-etal-2019-analyzing,geva2021whats}. Other approaches focus on the feed-forward layers \cite{geva2020transformer}. Gradient-based approaches, such as \cite{pmlr-v70-sundararajan17a} and \cite{smilkov2017smoothgrad} can be used to trace gradients, while influence functions trace model parameter changes throughout a LM \cite{10.5555/3305381.3305576,han-etal-2020-explaining}.
While backtracking all the model weights might be possible, such explanations can only tell us which part of the input they are looking at \cite{chen2019looks} and humans are ill-equipped to interpret them \cite{stammer2021right}. So instead, we aim for more intuitive and sparse explanations: well-descriptive but short-sequence prototypes. In addition to revealing parts of the input, this explanation type also shows prototypical cases similar to the input parts.

\begin{figure}[t]
    \small
	\centering
	\begin{center}
	\includegraphics[width=0.75\linewidth]{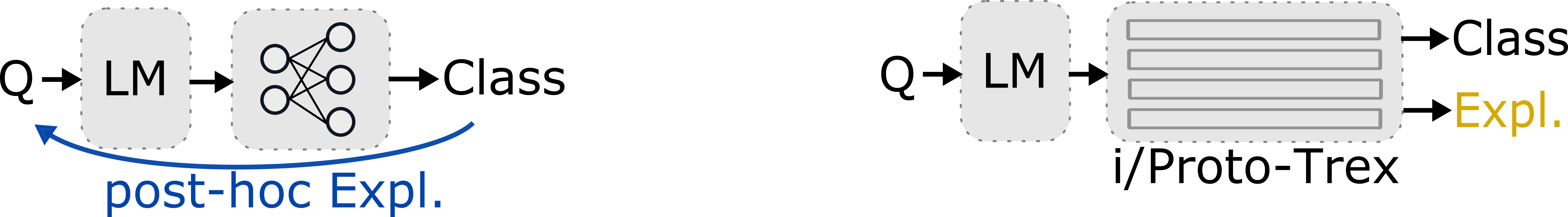}
	\end{center}
	{\def\arraystretch{1}\tabcolsep=2.pt
	\begin{tabularx}{\textwidth}{l|X}
		Explainer & Q: Staff is available to assist, but limited to the knowledge and understanding of the individual.\\
		\hline
        \multirow{2}{*}{Post hoc} & \colorbox{purple!20}{\makebox(13,6){Staff}}\colorbox{cyan!20}{\makebox(5,6){is}}\colorbox{cyan!35}{\makebox(29,6){available}}\colorbox{purple!20}{\makebox(5,6){\vspace{-1pt}to}}\colorbox{purple!10}{\makebox(20,6){\vspace{-1pt}assist,}}\colorbox{purple!40}{\makebox(7,6){but}}\colorbox{purple!15}{\makebox(22,6){limited}}\colorbox{purple!10}{\makebox(5,6){\vspace{-1pt}to}}\colorbox{cyan!15}{\makebox(8,6){\hspace{-2pt}the}}\colorbox{cyan!10}{\makebox(35,6){\vspace{-2pt}knowledge}}\colorbox{cyan!10}{\makebox(12,6){and}}\colorbox{cyan!20}{\makebox(50,6){\vspace{-2pt}understanding}}\colorbox{purple!5}{\makebox(5,6){of}} \colorbox{purple!5}{\makebox(7,6){the}}\colorbox{purple!60}{\makebox(35,6){individual.}}\\
        \hline
        Proto-Trex & They literally treat you like you are bothering them. No customer service skills.\\
        \hline
        iProto-Trex & They offer a bad service. \\
    \end{tabularx}
 	\caption{i/Proto-Trex compared to a post hoc explanation on sentiment classification. The input query (top row) is classified (as negative), and three explanations are provided. The first (post hoc) explanation is provided by LIT \cite{tenney2020language} with LIME \cite{ribeiro2016why}. The intensity of the color denotes the influence of a certain word: Blue (red) color indicates positive (negative) sentiment. Middle and bottom row explanations are provided by (interactive) Proto-Trex networks.}
 	\label{fig:motivation}
	}
\end{figure}

\paragraph{\bf Case-based Reasoning in Deep Neural Networks.}
Our work relates most closely to previous work demonstrating the benefits of prototype networks in Computer Vision \cite{chen2019looks, li2017deep} as well as for sequential data by combining them with RNNs \cite{hase2019interpretable, Ming_2019}.
More precisely, in contrast to post hoc interpretation methods, prototype networks employ explanations in the, e.g., classification process. That is, they classify a sample by comparing its parts --which could be words of a sentence-- to (learned) prototypical parts from other samples of a given class.
If things \textit{look} similar, they are classified similarly. Humans behave and reason very similarly, which is called case-based reasoning \cite{chen2019looks, caseCLIFTON2001765}. The prototypical parts provided by the network help the user understand the classification. 
Fig.~\ref{fig:motivation} illustrates the benefit of prototypical explanations.
One can observe that the post hoc explanation can leave the user clueless, while the prototypical explanations can help better understand the network's decision. Accordingly, prototypes are an additional method in the interpretability toolbox, extending current post hoc methods. 
In previous case-based approaches, however, the training of such networks was entirely data-driven. 
Therefore, like \cite{chen2019looks}, in our work, we do not focus on quantifying the interpretability unit of prototypical networks but extend the reasoning process of our network by case-based reasoning and Explanatory Interactive Learning (XIL) \cite{teso2019explanatory}. 

\paragraph{\bf Explanatory Interactive Learning.} 
The XIL framework \cite{teso2019explanatory} was proposed in order to promote richer communication between humans and machines, possibly to complement one another. More precisely, human users can ask the model to explain a specific prediction. They, in turn, can reverse the explanation flow and respond by correcting the model if necessary, providing \mbox{--not} necessarily \mbox{optimal--} feedback on the explanations \cite{stammer2021right}. 
The ultimate goal is to establish trust in the model not only by revealing a false reasoning process of the model through the model's explanations but also by giving the user the chance to correct the model through corrective feedback on these explanations. 
XIL showed in previous research to increase the performance and the explanation quality of deep black-box models and by that user's trust in them \cite{schramowski2020making, ross2017right, selvaraju2019taking}. 

We combine prototype networks and XIL into iProto-Trex. Prototypical explanations are ideally suited for user interactions compared to standard post hoc explanations as they allow the user to interact directly on the explanation , i.e. the prototypes, (\cf Fig~\ref{fig:3Interaction_pipeline}) instead of inserting the feedback indirectly, e.g., via the loss function.

\begin{figure*}
	\begin{subfigure}{0.48\columnwidth}
		\includegraphics[width=\linewidth]{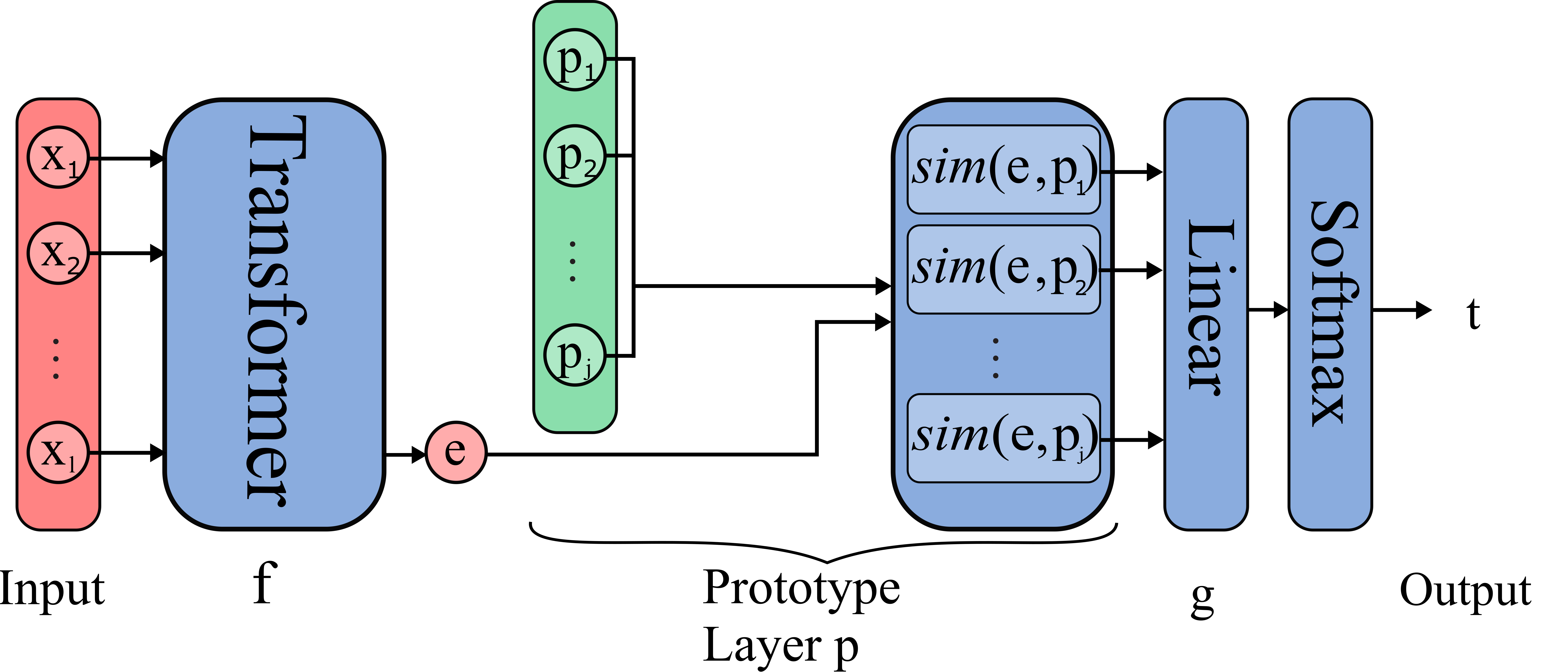}
		\caption{Sentence-Level}
	\end{subfigure}
	\begin{subfigure}{0.48\columnwidth}
    	\includegraphics[width=\linewidth]{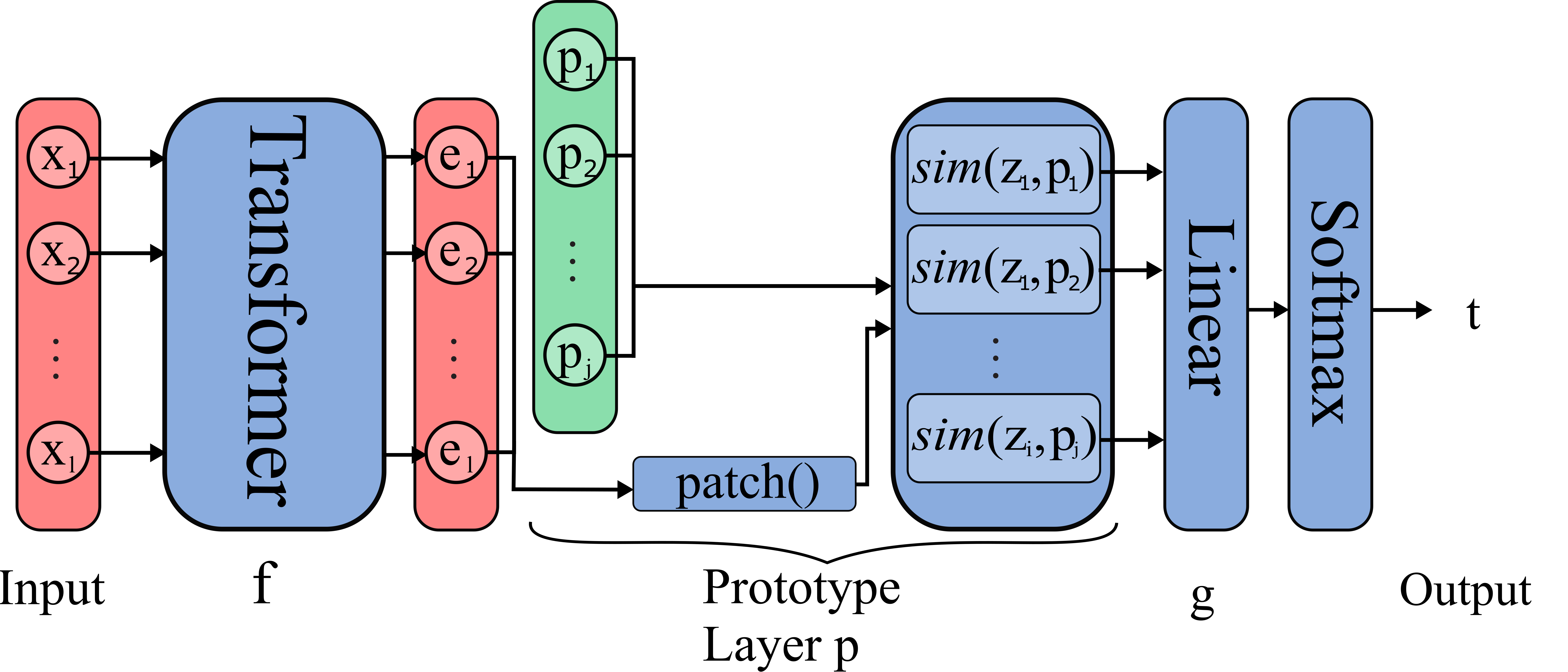}
		\caption{Word-Level}
	\end{subfigure}
	\caption{Architecture of the Prototypical-Transformer Explanation Network (Proto-Trex): (a) Sentence-level case uses a transformer to compute sentence-level embedding explanations, while (b) word-level case provides multiple word-level explanations and therefore requires an additional word-selection layer, \textit{patch()}.}
	\label{fig:3.2TProNet Architecture}
\end{figure*}

\section{Proto-Trex: Prototype Learning for Transformer LMs}
The prototype network (Proto-Trex) architecture builds on large-scale transformer LMs, summarized in Fig.~\ref{fig:3.2TProNet Architecture}. In the following, we describe the general idea of prototypical networks and our transformer-based prototype architecture and its modules in more detail.

\subsection{Proto-Trex Networks}
\paragraph{\bf General Idea.} Proto-Trex is a prototype network, where prototypes are representative observations in the training set. The classification is based on the neighboring prototypes, i.e. an input query is classified as positive when the network thinks it looks similar to a certain (near) positive prototype within the training set. 
In more detail, the reasoning and classification processes use the same module, i.e., they use the similarity between query and prototypes. As both reasoning and classification use the same module (in parallel), we consequently have an interpretable model.
In particular, (interpretable) prototypical explanations help verify Right for the Right Reasons \cite{ross2017right}.

\paragraph{\bf Architecture.} Specifically, the model, in our case the pre-trained transformer $f$, creates a context embedding $\mathbf{e}$ from the input sequence $\mathbf{x}$. This embedding contains useful features for the prediction and is then passed on to the prototype layer. This layer computes the similarity between the embedding $\mathbf{e}$ and each of the trainable prototypes $\mathbf{p}_j$ from the set of prototypes $\mathcal{P}$. The prototypes are learned during the training process and represent prototypical (sub-)sequences in the training data.
We use transformer models both on the sentence-level (a) and the word-level (b), resulting in a single input representation for the sentence-level transformer and $l$ for word-level. To compare the word-level embeddings $\mathbf{e}_i$ with the prototypes, we have to patch them ($\mathcal{Z}\!=\!\{patch(\mathbf{e}_i)\}$) into subsequences according to the desired prototype length $k$.
The resulting similarity values are passed on to the final linear layer $g$ with weight matrix $\mathbf{w}_g$ and no bias, mapping from $m$ (the number of prototypes) to the number of classes. Finally, the output values are passed on to a softmax producing class probabilities and predictions $t$.

\subsection{Proto-Trex Loss}
\label{subsec:optim}
Optimization of prototype networks aims at maximizing both performance and interpretability. To this end, our Proto-Trex loss $\mathcal{L}$ combines performance and interpretability with prior knowledge of the prototype network's structure. 
Let us illustrate this for the word-level case (the sentence-level case results from $\mathbf{e}\!=\!\mathbf{z}$): $\mathcal{L}\!\coloneqq\!$
\begin{align}
	\min_{\mathcal{P},\mathbf{w}_{g}} \frac{1}{n} \sum_{i=1}^{n} \text{CE}(t_i,y_i) 
	+ \lambda_1 \text{Clst}(\mathbf{z},\mathbf{p}) 
	+ \lambda_2 \text{Sep}(\mathbf{z},\mathbf{p}) 
	+ \lambda_3 \text{Distr}(\mathbf{z},\mathbf{p})
	+ \lambda_4 \text{Divers}(\mathbf{p})
	+ \lambda_5 ||{\mathbf{w}_{g}}|| \label{eq:4.2.loss_fct}
\end{align}
where $\lambda_i$ weights the influence of the different terms and $n$ is the number of training examples.
The first term is the cross-entropy (CE) loss optimizing the predictive power of the network. 
The second term (Clst) clusters the prototypes w.r.t.~the training examples of the same class, maximizing similarity to them (we rewrite maximization as minimization terms), and the separation loss minimizes the similarity to other-class instances:
\begin{equation}
	\text{Clst}(\mathbf{z},\mathbf{p})\coloneqq-\frac{1}{n} \sum_{i=1}^{n} \min_{j:\mathbf{p}_j\in\mathcal{P}_{y_i}} \min_{\mathbf{z}\in\mathcal{Z}} \text{sim}(\mathbf{z},\mathbf{p}_j) 	 \;;\;\; \text{Sep}(\mathbf{z},\mathbf{p})\coloneqq\frac{1}{n} \sum_{i=1}^{n} \min_{j:\mathbf{p}_j\notin\mathcal{P}_{y_i}} \min_{\mathbf{z}\in \mathcal{Z}} \text{sim}(\mathbf{z},\mathbf{p}_j)
	\label{eq:clst}
\end{equation}
Together, $\text{Clst}$ and $\text{Sep}$ push each prototype to focus more on training examples from the same class and less on training examples from other classes. Both are motivated by ProtoPNet \cite{chen2019looks}. To get prototypes that are distributed well in the embedding space, we introduce two additional losses, a distribution loss \cite{li2017deep}, assuring that a prototype is nearby each training example, and a diversity loss \cite{Ming_2019}:
\begin{equation}
	 \text{Distr}(\mathbf{z},\mathbf{p}) \coloneqq - \frac{1}{m} \sum_{j=1}^{m} \min_{i\in[1,n]} \min_{\mathbf{z}\in \mathcal{Z}} \text{sim}(\mathbf{z},\mathbf{p}_j) \;;\quad
 	 \text{Divers}(\mathbf{p}) \coloneqq \frac{1}{m} \sum_{\hat{\jmath}=1}^{m} \min_{j:\mathbf{p}_j\in\mathcal{P}}  \text{sim}(\mathbf{p}_{\hat{\jmath}},\mathbf{p}_j) \ 
	 \label{eq:4.2.distr_loss}
\end{equation}
In contrast to the other terms, the diversity loss does not compute similarities between embeddings and prototypes but between prototypes themselves. It is another way of distributing prototypes in the embedding space as it maximizes the distance between prototypes, preventing them from staying at the same --not necessarily optimal-- location.
This is especially helpful in the case of multiple prototypes for a single class, encouraging them to represent different facets of the class. If they are otherwise too similar, no information is gained, resulting in redundant prototypes -- together with the class-specific loss encouraged by the cluster loss ($\text{Clst}$) and the separation loss ($\text{Sep}$), this helps compute prototypes that focus solely on their class. Otherwise, we can get ambiguous prototypes leading to negative reasoning. Also, we clamp the weights of the classification layer with $min(\mathbf{w}_{g},0)$ to further avoid negative reasoning \cite{chen2019looks}.
Finally, the last term of the Proto-Trex loss (Eq.~\ref{eq:4.2.loss_fct}) is an L1 regularization term of the last layer $g$, which prevents the network from overfitting or relying too much on a single prototype. 

\subsection{Similarity Computation}\label{sec:3.2.3simi}
Computing similarities is an essential aspect of Proto-Trex. For prototypes to represent certain aspects or features of the input distribution in the embedding space, we compute the similarity \mbox{$\text{sim}(\mathbf{e},\mathbf{p})$} between an embedded training example and a prototype. We make use of \mbox{$\max \text{sim}(\mathbf{e},\mathbf{p})\!=\!\min \text{dist}(\mathbf{e},\mathbf{p})$} to replace each distance minimization with a similarity maximization.
We explore two approaches for the similarity computation:
\begin{align}
	\text{sim}(\mathbf{e},\mathbf{p}_j) = 
	\begin{cases}
		-\norm{\mathbf{e}-\mathbf{p}_j}_2 &,\;  \text{L2 norm }\\
		\frac{\mathbf{e} \cdot \mathbf{p}_j}{{||\mathbf{e}||}_2 ||\mathbf{p}_j||_2} &,\; \text{cosine similarity} \nonumber
	\end{cases}\;,
\end{align}
where the index $j$ denotes a specific prototype. For each training example, there is an embedding $\mathbf{e}$ which is compared to all $m$ prototypes $\mathbf{p}$, i.e. we get $m$ similarity values for each embedding. 
While the L2 norm computes the distance between two vectors, cosine similarity measures their angle. Both, but especially cosine similarity, are natural choices for NLP tasks \cite{IntroInfoRet}. 
However, cosine similarity is more robust here than the L2 norm as the distance's magnitude between vectors has no influence due to normalization.

\subsection{Selection for Word-level Prototypes}
Since learning prototypes for LMs pre-trained on word-level representations is more involved than for the sentence-level, let us focus on them here; the sentence-level case naturally follows from the discussion. Word-level prototypes are generally sensible as explanations should consist of sparse sequences, at best focusing only on subsequences of the input sentence. Moreover, as whole sequences can be ambiguous or contain little information, we combine different word embeddings ($patch()$) into subsequences and enforce the prototype to be similar to the relevant subsequences containing the most information. But how do we select the most informative words?

A naive approach would be to simply form all possible patches (word combinations) of the input sequence and compare them to the prototypes to find the best patch for the classification -- the most important subsequence should then be similar to a certain prototype. Unfortunately, for long input sequences with length $l$ it becomes hard to compute all possible word patches of length $k$, in this case $|\mathcal{Z}|\!=\!\binom{l}{k}$. To address this problem without losing too many valuable patches, the following two approaches are sensible ideas.

{\bf (a) Sliding windows} naturally reduce the number of patches. A sliding window is a convolution that selects a certain part of the input according to the window size and then sliding to the next part. The main disadvantage of sliding windows is the relatively rigid structure of a window. This is problematic in the context of NLP tasks that often have long-range dependencies. We introduce dilation to loosen this. Dilation facilitates looking at direct word neighborhoods but also at more distant ones to capture long-range dependencies. This ``convolutional'' approach is illustrated in Fig.~\ref{fig:word_proto_selection}(a). We note that applied to the contextualized word embeddings of transformer LMs, the convolutional approach without dilation should contain already global (long-range) information. Unfortunately, this information is stored in the embeddings and cannot be visualized easily.

{\captionsetup[figure]{skip=0pt}
\begin{figure}[t]
	\centering
	\begin{subfigure}[b]{0.49\columnwidth}
	    \begin{footnotesize}
    		\begin{tabularx}{\columnwidth}{cX}
                \multirow{2}{1pt}{\textbf{(1)}}&
                \colorbox{cyan!30}{The food}was delicious and the service great\\
                &$\;$The\colorbox{cyan!30}{food was}delicious and the service great\\\hline
                \multirow{2}{1pt}{\textbf{(2)}}&
                \colorbox{cyan!30}{The}food\colorbox{cyan!30}{\makebox(13,6){\vspace{-1.5pt}was}}delicious and the service great \\
                &$\;$The\colorbox{cyan!30}{food}was\colorbox{cyan!30}{delicious}and the service great
            \end{tabularx}
        \end{footnotesize}
        \caption{Sliding windows approach: Convolutions sliding over a sequence, (1) without and (2) with additional dilation in the convolutional window.}
	\end{subfigure} 
	\hspace{1pt}
	\begin{subfigure}[b]{0.49\columnwidth}
		\includegraphics[width=\linewidth]{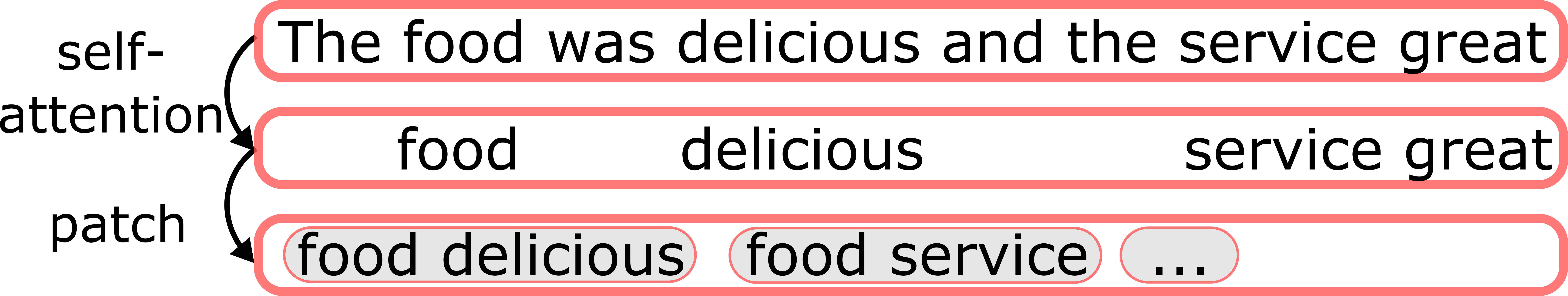}
		\caption{Self-Attention approach: First, the attention weights are calculated for each word, and most attended-to words are selected; then all patches w.r.t. the prototype length are determined.}
	\end{subfigure}
	\caption{Word selection for word-level Proto-Trex with (a) convolution and (b) attention.}
	\label{fig:word_proto_selection} 
\end{figure}}

As an alternative, the {\bf (b) Self-Attention} approach adds a self-attention layer after the transformer $f$ and before the prototype layer $p$ to filter irrelevant words (with low attention-scores), \cf Fig.~\ref{fig:word_proto_selection}(b). 
The purpose of this self-attention layer differs from the one in the transformer LM itself. The attention mechanism used here selects only the most important words, and the embedding representation remains untouched. To still provide as much information and variety as possible, the number of selected words $n_w$ of the attention layer, passed into the distance computation, is a hyperparameter, and we chose it to be twice as much as the length of a single prototype; however, clamped by the threshold $k_{\text{lim}}$, 
$n_w \!=\! min(k_{\text{lim}}, \,2k)$.
The threshold can be set w.r.t.~computational efficiency. 
This means, e.g., having a desired prototype length of $k\!=\!4$ with $k_{\text{lim}}\!=\!10$, the attention layer selects $n_w\!=\!8$ words yielding in this example $|\mathcal{Z}|\!=\!\binom{8}{4}\!=\!70$ patch combinations for the distance computation. 

\subsection{Decoding via Nearest Neighbor Projection} \label{sec:nn_projection}
Proto-Trex networks encode prototypes in the embedding space. Consequently, they cannot simply be decoded from the transformer embedding space, as this space and textual data are categorical and not continuous. To overcome this, we assign, i.e. project, each prototype to its nearest neighbor from the training data (where the textual representation is available) in an intermediate training step.
Thereby, we ensure that the prototype really represents what it actually looks like, which also increases the interpretability of the Proto-Trex network. 
The final training step (after the nearest neighbor projection) then fine-tunes the classification head to adapt it to the projected prototypes.
Doing so has the advantage of being decoded precisely on the spot and giving more certainty about the explanation provided. The disadvantage, however, is that a prototype may be located sub-optimally, which could lead to performance losses.
We faced this issue especially for word-level prototypes and found the projection ineligible for these highly contextualized word representations.
Instead, for word-level, we only use the nearest neighbor for approximating the learned prototype in order to provide the explanation.

\section{Interactive Prototype Learning}
Recent case-based reasoning approaches in Computer Vision and NLP usually assume a large volume of data for training \cite{chen2019looks, li2017deep, hase2019interpretable}, with little or no user feedback during the model building process. However, user knowledge and interaction, in particular via explanations, can be valuable already in the model building and understanding phase, significantly reducing the amount of data required, avoiding Clever-Hans moments early on, increasing the explanation quality of the model and, in turn, user trust \cite{schramowski2020making,teso2019explanatory}.

\begin{figure}[t]
	\centering
	\includegraphics[width=0.9\columnwidth]{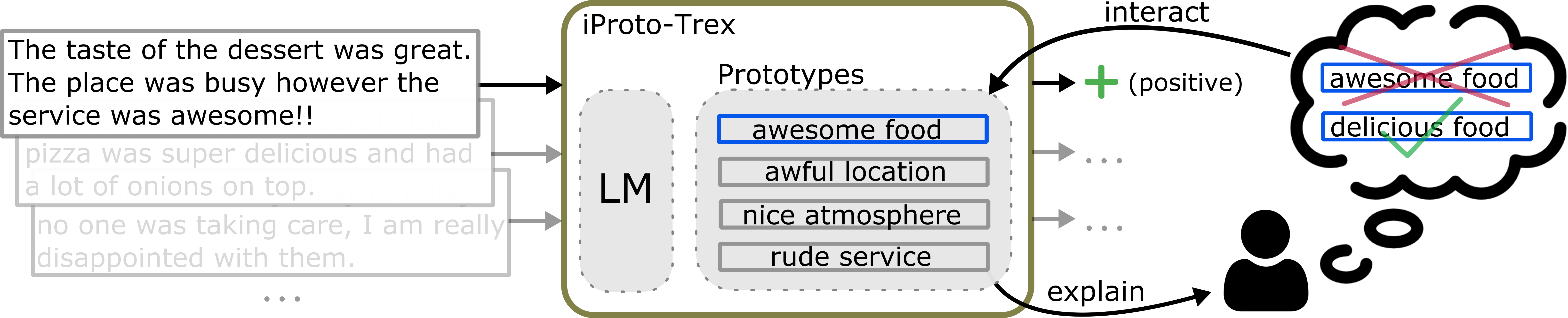}
	\caption{Interactive Prototype Learning: iProto-Trex classifies the input and gives the user an explanation based on a prototype. The explanation is highlighted in bright blue. If the user is dissatisfied with the given explanation, they can replace it with a self-chosen sequence.
	}
	\label{fig:3Interaction_pipeline}
\end{figure}

\paragraph{\bf Interactive Proto-Trex.}
To address this, we propose an explanatory interactive learning approach for Proto-Trex networks, called iProto-Trex, as illustrated in Fig.~\ref{fig:3Interaction_pipeline}. 
Specifically, carried out during the training progress by freezing and evaluating the current state or post hoc training, the interaction takes the following form. At each step, Proto-Trex networks provide prototypes as explanations of a classification. The user responds by correcting the learner, if necessary, in our case, the provided prototypes. For this, iProto-Trex provides several options to interact with prototypes. 

Specifically, iProto-Trex distinguishes between weak- and strong-knowledge interactions. In strong-knowledge interactions, users are certain about their feedback (which could require high-level expertise). In this case, iProto-Trex offers to \textit{remove}, \textit{add} or \textit{replace} prototypes to manipulate the explanation according to user preferences. While removing a prototype helps if the network has learned redundant prototypical sequences or a subset already covers all critical aspects of the task at hand, adding new prototypes considers user preferences and knowledge.
Replacing a prototype is the same as removing plus adding.
After these interactions, we retrain the subsequent classification layer to adapt the network to the updated prototype setup.

In weak-knowledge interactions, users only state (dis)satisfaction with prototypes based on their intuition. They must not know what a replacement should look like, which is a good trade-off between user knowledge and loss optimization of the network.
To this end, iProto-Trex offers \textit{re-initialization} and \textit{fine-tuning}; instead of providing an explicit replacement, users freeze prototypes they like while relearning those they are dissatisfied with. 
The difference between these approaches is that users can express how well the current prototype represents a particular task aspect. Another form of weak-knowledge interaction for sentence-level prototypes is \textit{pruning} the sequence length of prototypical explanations, i.e., compressing it to the essentials. In order to limit meaning changes of a sequence, a threshold is provided for the cosine similarity of the pruned and the original version. We initially set this threshold to $0.8$, but users can set the threshold as they like depending on the certainty of their knowledge. 

\paragraph{\bf Soft User Feedback.}
iProto-Trex's interaction methods require users to be quite certain about their feedback, especially in the case of strong-knowledge interactions, --they have to be experts.
To elevate this burden, we propose a \textit{soft feedback} mechanism, using a loss based on the prototype similarity as
\begin{align}
	\label{eq:interactLoss}
	\mathcal{L}_{\text{interact}}\coloneqq \lambda_6\max_{\mathbf{p}_{\text{old}}} \big( max(\text{sim}(\mathbf{p}_{\text{old}}, \mathbf{p}_{\text{new}}),c) \big)
\end{align}
with $c\!\in\![0,1]$. Instead of directly replacing the selected prototype $\mathbf{p}_{\text{old}}$, this soft interaction loss pushes it to be similar to the user-suggested prototype $\mathbf{p}_{\text{new}}$. Moreover, users can control how strong iProto-Trex should incorporate their feedback by setting the certainty value $c$ between $0$ (low certainty) and $1$ (high certainty). 

\section{Faithful Prototypical Explanations}
A question that naturally arises when dealing with explainable AI is how to evaluate the quality of an explanation. One of the key metrics in NLP and Computer Vision is the faithfulness of explanations \cite{faithfulattribution,deyoung-etal-2020-eraser}, i.e. whether a given explanation represents the true reasoning process of the black-box model. 
First, we use the popular benchmark of \cite{deyoung-etal-2020-eraser} and evaluate the faithfulness of Proto-Trex as a whole by showing that the class probabilities change significantly if we perturb the input. 
We follow their approach and compute comprehensiveness and sufficiency by removing the rationale in each input sequence and evaluating the changes in the class probabilities for the prediction.
Second, we perturb the prototype layer by removing the top explanation for a given test sample and compute the loss in accuracy. This helps identify to what extent the decision was actually based on the explanation given. Since our prototypical explanations are already short sequences, removing the entire prototype is similar to removing the rationale in the explanation.

\begin{table}[t]
    \small
	\parbox{.45\linewidth}{
	{\def\arraystretch{1.}\tabcolsep=2.pt
	\centering
	\begin{tabular}{c|l|c|c|c}
		\textbf{a)} &Language Model & Yelp & Movie & Toxic \\
		\hline
		 \parbox[t]{2mm}{\multirow{4}{*}{\rotatebox[origin=c]{90}{sent.-level}}}&SBERT & $94.92$ & $84.56$ & $\circ\mathbf{84.40}$ \\
		&SBERT (ours) & $93.59$ & $80.05$ & $73.19$ \\
		&CLIP & $93.78$ & $75.49$ & $80.82$ \\
		&CLIP (ours) & $87.16$ & $63.52$ & $67.75$ \\
		\hline
		 \parbox[t]{2mm}{\multirow{6}{*}{\rotatebox[origin=c]{90}{word-level}}}&BERT & $89.41$ & $61.45$ & $76.88$ \\
		&BERT (ours)& $92.10$ & $75.51$ & $79.35$\\
		&GPT-2  & $93.78$ & $\bullet\mathbf{87.05}$ & $81.56$\\
		&GPT-2 (ours)& $\bullet\mathbf{95.32}$ & $84.57$ & $84.14$\\
		&DistilBERT  & $92.91$ & $79.62$ & $81.57$ \\
		&DistilBERT (ours) & $92.71$ & $78.64$ & $83.39$\\
		\hline
		 \parbox[t]{2mm}{\multirow{2}{*}{\rotatebox[origin=c]{90}{inter.}}}&SBERT (ours) & $93.81$ & $80.24$ & $73.36$ \\
		&GPT-2 (ours) & $\circ\mathbf{95.25}$ & $\circ\mathbf{84.80}$ & $\bullet\mathbf{84.51}$ \\
	\end{tabular}
	}
	}
	\hspace{12pt}
	\parbox{.49\linewidth}{
	{\def\arraystretch{1.}\tabcolsep=2.pt
	\begin{tabular}{l|cc|cc}
		\textbf{b)} & \multicolumn{2}{c|}{Word-Selection} & \multicolumn{2}{c}{Similarity} \\
		\hline
		Proto-Trex LM & Conv. & Attn. & Cos. & L2 \\
		\hline
		BERT & $\mathbf{92.10}$ & $89.66$ & $\mathbf{92.10}$ & $91.34$\\
		GPT-2 & $\mathbf{95.32}$ & $94.16$ & $\mathbf{95.32}$ & $94.99$\\ 
		DistilBERT & $\mathbf{92.71}$ & $91.09$ &$\mathbf{92.71}$& $92.05$\\ 
		\hline
		SBERT & -- & -- & $\mathbf{93.59}$ & $93.13$ \\ 
		CLIP & --& -- & $\mathbf{87.16}$ & $86.88$\\
	\end{tabular}
	}
    
    {\def\arraystretch{1.}\tabcolsep=2.pt
	\begin{tabular}{l|cc|cc}
		\textbf{c)} & \multicolumn{2}{c|}{Faithfulness} & \multicolumn{2}{c}{Acc.} \\
		\hline
		Proto-Trex LM & Comp. & Suff. & before & after \\
		\hline
		SBERT & $0.22$ & -$0.08$ & $80.05$ & $69.66$\\
		iSBERT & $0.21$ & -$0.08$ & $80.24$ & $70.76$\\ 
		\hline
		GPT-2 & $0.12$ & $0.02$ & $84.57$ & $49.91$ \\
		iGPT-2 & $0.13$ & $0.02$ & $84.80$ & $55.74$ \\ 
	\end{tabular}
	}
	}
	\caption{Quantitative Results. a) Average Accuracy of Proto-Trex with different LMs on different datasets. Best (``$\bullet$'') and runner-up (``$\circ$'') are \textbf{bold}.	b) Ablation study of Proto-Trex module choices on Yelp dataset. c) Evaluation of i/Proto-Trex regarding faithfulness on MovieReview dataset. Mean values over $5$ runs are reported. The confidence intervals can be found in Appendix Tab.~\ref{tab:exp_performance_appendix}}
	\label{tab:exp_performance}
\end{table}

\section{Experimental Evaluation}
Our intention here is to investigate how good prototypes help understand transformer LMs. To this end, we evaluated i/Proto-Trex explanations on three benchmark datasets: MovieReview \cite{moviereviewdata2002}, Open Yelp\footnote{https://www.yelp.com/dataset}
and Jigsaw Toxicity\footnote{https://www.kaggle.com/c/jigsaw-toxic-comment-classification-challenge}. We compared five pre-trained LMs (GPT-2 \cite{radford2019language}, BERT \cite{devlin2019bert}, DistilBERT, \cite{sanh2020distilbert}, SBERT \cite{reimers2019sentencebert} and the text-encoder of CLIP \cite{radford2021learning}) to investigate three questions: 
\begin{description}
\item[(Q1)] How much does adding a prototype layer affect the performance of (non-interpretable) LMs, i.e. a classification head defined by two fully-connected non-linear layers? 
\item[(Q2)] How does the performance change after and during interaction with the model explanations? In particular, we investigated the different modules of Proto-Trex networks on sentence- and word-level and the interaction between users and the prototypical explanations.  
\item[(Q3)] How faithful are the given explanations, i.e. how well does the given explanation represent the true reasoning of the black-box model?
\end{description}
We present qualitative and quantitative results and refer to the Appendix for additional details on the experiments and our implementation, as well as additional qualitative results.
If not stated otherwise, the Proto-Trex architecture includes sentence- and word-level embeddings, the convolution module without dilation to select word-tokens, and cosine similarity to compute the similarity between prototypical explanations and input query.
We optimized the prototype and classification module with the proposed loss (Eq.~\ref{eq:4.2.loss_fct}) and initialize the prototypes randomly.
In each experiment, the number of prototypes is $m\!=\!10$, and the largest variant of the corresponding pre-trained LM is evaluated.

\paragraph{\bf (Q1,2) Trade off Accuracy and Interpretability.}
Tab.~\ref{tab:exp_performance} summarizes the experimental result of Proto-Trex network based on different sentence- and word-level LMs.  
As one can see and expected from the literature, interpretability comes along with a trade-off in accuracy.
The trade-off is generally higher on sentence-level LMs, partially due to the nearest neighbor projection (\cf Appendix for direct comparison).
However, the difference between traditional LMs and Proto-Trex LMs is often marginal and task-dependent (e.g. DistilBERT on Yelp and Movie).
Surprisingly, in the case of BERT and GPT-2 (on Yelp and Toxicity), the Proto-Trex network is outperforming the baseline LMs. 
Most interestingly, one can observe that the user may boost the performance of the corresponding Proto-Trex network interactively (iProto-Trex). Overall, i/Proto-Trex is competitive with state-of-the-art LMs while being much more transparent.   

\begin{table*}[t]
    \small
	\centering
	{\def\arraystretch{1.5}\tabcolsep=2.pt
	\begin{tabularx}{\textwidth}{cc|X}
	    & Importance & Query: I really like the good food and kind service here.\\
		\hline
		\parbox[t]{3mm}{\multirow{7}{*}{\rotatebox[origin=c]{90}{sentence-level}}}&${\scriptstyle0.52\cdot8.07=}\,\mathbf{4.20}$ & \textit{P8}: Oliver Rocks! Great hidden gem in the middle of Mandalay! Great friends great times\\
		&\multirow{2}{*}{${\scriptstyle0.84\cdot3.04=}\,\mathbf{2.55}$} & \textit{P4}: Incredibly delicious!!! Service was great and food was awesome. Will definitely come back!\\
        &${\scriptstyle0.90 \cdot1.21=}\,\mathbf{1.10}$ & 
        \setlength{\fboxrule}{2pt}\fcolorbox{violet}{white}{\parbox{0.79\columnwidth}{\textit{P6}: I found this place very inviting and welcoming. The food is great so as the servers.. love the food and the service is phenomenal!! Well done everyone..!}}
        \\
		& ${\scriptstyle0.79 \cdot 1.16 =}\, \mathbf{0.92}$ & \setlength{\fboxrule}{2pt}\fcolorbox{cyan}{white}{\parbox{0.79\columnwidth}{\textit{P2}: This place is really high quality and the service is amazing and awesome! Jayden was very helpful and was prompt and attentive. Will come back as the quality is so good and the service made it!}} \\
		\hline
		\parbox[t]{3mm}{\multirow{5}{*}{\rotatebox[origin=c]{90}{after interaction}}}&\multirow{2}{*}{${\scriptstyle0.84\cdot5.02=}\,\mathbf{4.22}$} & \textit{P4}: Incredibly delicious!!! Service was great and food was awesome. Will definitely come back!\\
		&${\scriptstyle0.92 \cdot2.10=}\,\mathbf{1.94}$ & 
		\setlength{\fboxrule}{2pt}\fcolorbox{violet}{white}{\parbox{0.79\columnwidth}{\textit{P6}: I found this place very inviting and welcoming. The food is great so as the servers.}}\\
		&${\scriptstyle0.76\cdot2.13=}\,\mathbf{1.62}$ & \setlength{\fboxrule}{2pt}\fcolorbox{cyan}{white}{\parbox{0.79\columnwidth}{\textit{P2}: This place is really high quality and the service is amazing and awesome!}}\\
		&${\scriptstyle0.73\cdot0.71=}\,\mathbf{0.52}$ & \textit{P10}: Fun and friendly atmosphere, fantastic selection. The sushi is so fresh\\
	\end{tabularx}
	}
	\caption{Provided Explanations by Proto-Trex networks. 
	Top-four explanations for the query (top row) are provided for sentence-level networks. Colored boxes illustrate the advantage of pruning sentence explanations (interaction). Importance scores (left, bold) are calculated with cosine similarity and classification weight.
	}
	\label{tab:exp_query_word}
\end{table*}

\paragraph{\bf (Q1) Ablation Study.}
Next, we investigate different Proto-Trex module choices for performance, \cf Tab.~\ref{tab:exp_performance}b), and explanation provision impact. While the input-prototype similarity computation does not explicitly correlate with the prototypes learned --yet we assumed it to be the better choice--, the word-selection module also impacts the explanation outcome. Furthermore, cosine similarity is not only more accurate but also converges faster. 
In terms of accuracy, the convolutional word selection outperforms the attention module. Moreover, we observe an advantage for the provided explanations; namely, the attention module tends to select punctuation and stop words. According to \cite{ethayarajh2019contextual}, this is because punctuation and stop-words are among the most context-specific word representations--they are not polysemous but have an infinite number of possible contexts.
The convolution module, in contrast, is easier to interpret as words are coherent. 

\begin{table*}[t]
    \small
    \centering
    {
	\begin{tabularx}{\textwidth}{c|c|X}
		Type of interaction &  Acc. & Prototype \\
        \hline
		\multirow{2}{*}{no interaction} & \multirow{2}{*}{$\scriptstyle93.64$} & Horrible customer service and service does not care about safety features. That's all I'm going to say. Oh they also don't care about their customers\\
		\hline
		\multirow{3}{*}{re-initialize} & 
		\multirow{3}{*}{$\scriptstyle93.80$} & Terrible delivery service. People are mean and don't care about their customers service. I will not ever come back to this place. Also the food is small, not very good and too expensive. \\\hline
		\multirow{3}{*}{soft replace ($0.5$)}& \multirow{3}{*}{$\scriptstyle93.81$}& Terrible delivery service. People are mean and don't care about their customers service. I will not ever come back to this place. Also the food is small, not very good and too expensive. \\\hline
		\multirow{2}{*}{soft replace ($0.9$)}& \multirow{2}{*}{$\scriptstyle93.79$} & I really don't recommend this place. The food is not good, service is bad. The entertainment is so cheesy. Not good \\\hline
		soft replace ($1.0$) & {$\scriptstyle93.79$} & They offer a bad service. \\
	\end{tabularx}
	}
	\caption{Interactive learning: Different user interaction methods with accuracy on Yelp reviews. 
	Interaction changes a redundant prototype into a user-selected one incorporating user preferences and their notion of a \textit{good} explanation without significantly impacting the accuracy.
	}
	\label{tab:exp_interaction}
\end{table*}

\paragraph{\bf (Q1,2) Provided Explanations.}
To analyze the provided explanations, we consider the SBERT based Proto-Trex networks, \cf Tab.~\ref{tab:exp_performance}, trained on the Open Yelp dataset. 
Tab.~\ref{tab:exp_query_word} shows how Proto-Trex provides users with explanations. These explanations correspond to prototypical sequences for a query. In addition, Proto-Trex provides corresponding importance scores, indicating the significance of an explanation for the classification. We show four sentence-level prototypes for the query so users can quickly extract the essential aspects that help understand the classification. However, one can observe that the sentence-level explanations sometimes lack sparsity and are difficult to interpret w.r.t. the query, demonstrating the demand for (further) interaction (\cf Tab.~\ref{tab:exp_query_word}, colored boxes with prototypical explanations \textit{P6} and \textit{P2}). The pruned Proto-Trex provides less ambiguous and easier to interpret sequences as prototypical explanations.


\paragraph{\bf (Q2) Interactive Prototype Learning.}
In the previous experiment, pruning has already shown the benefits of adapting explanations to user preferences.
In order to further investigate our interactive learning setting, we incorporated certainty ($c$) and examined its effectiveness.
Since sentence-level prototypes allow for better coherence, we will focus on them in the interactive learning.
Tab.~\ref{tab:exp_interaction} shows an influential, i.e. high importance value, sentence-level prototype (yet no interaction) for negative restaurant reviews on the Open Yelp dataset.
Assuming a user is dissatisfied with the prototypical explanation yet uncertain about what a good explanation would entail: With re-initialization (most uncertain interaction technique), we can already observe that users can influence the network's decision process based on their intuition of a \textit{weak} component without performance loss.
However, the revised explanation is still not sufficiently interpretable, i.e. too long. Therefore, we considered incorporating explicit user feedback here. In this case, the phrase ``They offer a bad service'' served as soft replacement for the model's provided explanation, and the user applied it with different levels of certainty. First, a low certainty value (\mbox{$c\!=\!0.5$}) results in the same explanation as before. This is because the similarity of any two prototypes with clearly negative sentiment is higher than a certainty threshold of $0.5$. In this way, the model does not ``stubbornly'' adopt the feedback but trades off the user certainty and the model performance. As the user gets more certain, he gradually increases the threshold. Finally (\mbox{$c\!=\!1$}), the user simply replaced the prototype to obtain their desired solution, again without compromising accuracy. From this interactive conversation, we obtain a bidirectional interaction loop between humans and AI, representing a more daily human-to-human discussion. The loop can be repeated multiple times with different feedback types or certainty values until the user is satisfied.
In summary, our results demonstrate that explanatory interactive learning is a powerful bidirectional method to adapt the network according to user preferences of \textit{good} prototypical explanations along with high accuracy. 

\paragraph{\bf (Q3) Faithful Prototypes.}
To analyze faithfulness, we first follow \cite{deyoung-etal-2020-eraser} and use the MovieReview dataset as they provide human-annotated rationales for this dataset. We compare the classification probabilities of the samples with the samples where the rationale is removed (comprehensiveness) or with only the rationale (sufficiency). Tab.~\ref{tab:exp_performance}(c) shows that i/Proto-Trex scores high for comprehensiveness, indicating that the network is focusing on the input rationale for the classification, while low sufficiency scores indicate that the surrounding context has little to even poor impact on the prediction. This shows that i/Proto-Trex are generally faithful classifiers, i.e. the human-annotated rationales agree with the internal rationales in the model. 
Furthermore, we remove the top (prototypical) explanation for the classification and examine the change in accuracy after the removal to show that it is the prototype that faithfully captures the rationale. Tab.~\ref{tab:exp_performance}(c) shows a drastic decrease in accuracy for both models after the explanation removal, confirming the faithful contribution of the prototypes to the classification.

\section{Ethics Statement}
In addition to the previous sections' already included ethical considerations, we want to emphasize further ethical aspects here. This work approaches the explainability of transformer language models through case-based reasoning in the form of prototypical explanations. The interpretable prototype module builds on a pre-trained transformer language model. As these models are trained on data that is not publicly available, it remains unclear whether any kind of bias is inherent to the whole model \cite{bender2021stochasticparrots__}. In addition, concerns about privacy violations and other potential misuse emerge as they are trained with weak supervision. XAI inherently suffers from reporting bias as the presented explanations offer a vast space for over- and misinterpretation. Prototype networks and thus our proposed method significantly attenuate this problem but cannot entirely remove it.

\section{Conclusion}
Large-scale transformer LMs, like other black-box models, lack interpretability. We presented methods (prototype networks) to incorporate case-based reasoning to explain the LM's decisions. Despite the explanatory power of prototypical explanations, challenges regarding the quality of their interpretability still exist. Previous applications lack human supervision, although case-based reasoning is a human-inspired approach. Therefore, we propose an interactive prototype learning setting to overcome these challenges and improve the network's capabilities by incorporating human knowledge with the consideration of knowledge certainty. 
An exciting future avenue is to equip prototype networks with a more flexible interaction policy, i.e. components beyond user certainty, to promote a greater human-AI communication towards what might be called cooperative AI~\cite{Dafoe2021nature}.

\bibliography{references}
\bibliographystyle{vancouver}

\clearpage

\appendix
\section{Appendix}

\subsection{Datasets}
The benchmark datasets mentioned above provide a fixed test set except the Yelp Open dataset. For the Yelp Open dataset we randomly select $200\,000$ training examples and split them into train ($70\%$), validation ($15\%$) and test ($15\%$) set.  We split the Jigsaw Toxicity train set into train ($20\%$) and validation ($80\%$) set. Before the split, we filter out long sequences (\mbox{$num\_tokens\!>\!40$}) for each dataset. This is required because transformer-based LMs can only handle sequences of limited length, especially CLIP, and long sequences also cause the other sequences of a set to be padded to the same length, which, in turn, produces many padding tokens. For the tokenization we use the GPT2-tokenizer\footnote{\url{https://huggingface.co/transformers/model\_doc/gpt2.html\#gpt2tokenizer}}. 
We apply grid search for hyperparameters optimization. We report the cross-validated results. For the faithfulness computation we use the \textit{eraserbenchmark} repository\footnote{\url{https://github.com/jayded/eraserbenchmark}} and data\footnote{\url{https://www.eraserbenchmark.com/}}.

\subsection{Training}
The Proto-Trex networks are trained with and evaluated on different datasets. We use PyTorch\footnote{\url{https://pytorch.org/}} for the implementation. We optimize our model with Adam optimizer with hyperparameters \mbox{$\boldsymbol{\beta}\!=\!(0.9,0.999)$} and \mbox{$\epsilon\!=\!10^{-8}$}. We use a base learning rate of \mbox{$lr_{\text{base}} \!=\! 0.001$} and apply learning rate warm-up and scheduling that is here a linear decay. 
The learning rate is then given as 
\begin{align}
	lr = lr_{\text{base}} \cdot min
	\left(\frac{step_{i}}{e_{\text{wup}}},
	\frac{e - step_{i}}{e - e_{\text{wup}}}
	\right) \ , \nonumber
\end{align}
where $step_i$ is the current step, $e$ the total number of epochs and $e_{\text{wup}}$ the number of warm-up epochs.
The warm-up takes place for $e_{\text{wup}}\!=\!min\left(10, \frac{e}{20}\right)$ epochs. Warm-up reduces the dependency of early optimization steps that may cause difficulties in the longer run. Also, we weigh the cross-entropy loss with the class occurrences to re-balance the influence of an unbalanced dataset. The same is done in the accuracy computation where a balanced class accuracy is computed. For regularization, we apply an L1 regularization on the weights of the last linear layer. During the training process, a network may tend to overfit. To counteract this, we evaluate the model every $10^{\text{th}}$ epoch on the validation set and keep the model that yielded the best validation result. Furthermore, we set up a class mask to assign each class to the same number of prototypes. This is used and enforced by the class-specific losses (Clst and Sep). Changing the balance of the class assignment can be sensible to correct for imbalance in the dataset or focus on a specific class if its sentiment is more polysemous or generally more relevant. We show the hyperparameters used for our experiments, found by a grid search. The found hyperparameters were often very similar across models and datasets as the parameters mainly aimed at interpretability, not only performance. Note, not all hyperparameters were grid-searched simultaneously. We first searched for the best combination of $\lambda_1$ to $\lambda_5$ and searched for $\lambda_6$ separately because the interaction loss depends on interaction and does not directly impact the model performance during training. The best performing value for $\lambda_6$ was $0.5$. For the attention-based word selection we choose the number of heads to be $1$ and $k\!=\!4;k_{lim}\!=\!10$.
\begin{table}[t]
    \centering
    \begin{tabular}{c|c|c|c|c|c}
        \multicolumn{1}{l|}{\textbf{1)} \hspace{0.6cm} $\lambda_1$} & $\lambda_2$ & $\lambda_3$ & $\lambda_4$ & $\lambda_5$ & $\lambda_6$\\
        \hline
        $\{0, 0.1, 0.2, 0.5\}$ & $\{0, 0.1, 0.2, 0.5\}$ & $\{0, 0.1, 0.2\}$ & $\{0, 0.1, 0.3\}$ & $\{0, 0.001\}$ & $\{0, 0.1, 0.5, 1.0\}$
    \end{tabular}
	\newline
    \vspace*{1mm}
    \newline
    \begin{tabular}{c|c|c|c|c|c|c}
        \textbf{2)} & LM & $\lambda_1$ & $\lambda_2$ & $\lambda_3$ & $\lambda_4$ & $\lambda_5$ \\
        \hline
        \parbox[t]{2mm}{\multirow{5}{*}{\rotatebox[origin=c]{90}{Movie}}} & SBERT & $0.5$ & $0.2$ & $0.2$ & $0.3$ & $0.001$ \\ 
        & CLIP & $0.5$ & $0.2$ & $0.2$ & $0.3$ & $0.001$ \\ 
        & BERT & $0.2$ & $0.2$ & $0.2$ & $0.3$ & $0.001$ \\ 
        & GPT-2 & $0.2$ & $0.2$ & $0.1$ & $0.3$ & $0.001$ \\ 
        & DistilBERT & $0.2$ & $0.2$ & $0.1$ & $0.3$ & $0.001$ \\ 
    \end{tabular}
	\newline
    \vspace*{1mm}
    \newline
    \begin{tabular}{c|c|c|c|c|c|c}
        \textbf{3)} & LM & $\lambda_1$ & $\lambda_2$ & $\lambda_3$ & $\lambda_4$ & $\lambda_5$ \\
        \hline
        \parbox[t]{2mm}{\multirow{5}{*}{\rotatebox[origin=c]{90}{Yelp}}} & SBERT & $0.5$ & $0.2$ & $0.1$ & $0.3$ & $0.001$ \\ 
        & CLIP & $0.5$ & $0.2$ & $0.2$ & $0.3$ & $0.001$ \\ 
        & BERT & $0.5$ & $0.2$ & $0.1$ & $0.3$ & $0.001$ \\ 
        & GPT-2 & $0.5$ & $0.2$ & $0.2$ & $0.3$ & $0.001$ \\ 
        & DistilBERT & $0.5$ & $0.2$ & $0.2$ & $0.3$ & $0.001$ \\ 
    \end{tabular}
	\newline
    \vspace*{1mm}
    \newline
    \begin{tabular}{c|c|c|c|c|c|c}
        \textbf{4)} & LM & $\lambda_1$ & $\lambda_2$ & $\lambda_3$ & $\lambda_4$ & $\lambda_5$ \\
        \hline
        \parbox[t]{2mm}{\multirow{5}{*}{\rotatebox[origin=c]{90}{Toxicity}}} & SBERT & $0.2$ & $0.1$ & $0.1$ & $0.1$ & $0.001$ \\ 
        & CLIP & $0.2$ & $0.1$ & $0.2$ & $0.1$ & $0.001$ \\ 
        & BERT & $0.2$ & $0.2$ & $0.2$ & $0.1$ & $0.001$ \\ 
        & GPT-2 & $0.2$ & $0.2$ & $0.1$ & $0.1$ & $0.001$ \\ 
        & DistilBERT & $0.2$ & $0.2$ & $0.1$ & $0.1$ & $0.001$ \\ 
    \end{tabular}
	\newline
    \vspace*{1mm}
    \newline
    \caption{1) Search space for hyperparameter grid search. 2-4) Hyperparameter setup for Proto-Trex models on the three different datasets.}
    \label{tab:my_label}
\end{table}

\subsection{Sentence- vs. Word-level Loss}
For clarification, we present the loss terms for sentence-level i/Proto-Trex networks in more detail: $\mathcal{L}\!\coloneqq\!$
\begin{align}
	\min_{\mathbf{P},\mathbf{w}_{g}} \frac{1}{n} &\sum_{i=1}^{n} \text{CE}(t_i,y_i) 
	+ \lambda_1 \text{Clst}(\mathbf{e},\mathbf{p}) 
	+ \lambda_2 \text{Sep}(\mathbf{e},\mathbf{p}) 
	+ \lambda_3 \text{Distr}(\mathbf{e},\mathbf{p})
	+ \lambda_4 \text{Divers}(\mathbf{p})
	+ \lambda_5 ||{\mathbf{w}_{g}}|| \;, \label{eq:4.2.loss_fct_appendix}
\end{align}
where 
\begin{align}
    \text{Clst}(\mathbf{e},\mathbf{p}) &\coloneqq
	-\frac{1}{n} \sum_{i=1}^{n} \min_{j:\mathbf{p}_j\in\mathbf{P}_{y_i}} \text{sim}(\mathbf{e}_i,\mathbf{p}_j) \; , \\
    \text{Sep}(\mathbf{e},\mathbf{p}) &\coloneqq
	 \frac{1}{n} \sum_{i=1}^{n} \min_{j:\mathbf{p}_j\notin\mathbf{P}_{y_i}} \text{sim}(\mathbf{e}_i,\mathbf{p}_j)\; , \\
    \text{Distr}(\mathbf{e},\mathbf{p}) &\coloneqq
	 - \frac{1}{n} \sum_{j=1}^{n} \min_{i\in[1,n]} \text{sim}(\mathbf{e}_i,\mathbf{p}_j) \ , \\ \intertext{and}
    \text{Divers}(\mathbf{p}) &\coloneqq
	 \frac{1}{m} \sum_{\hat{\jmath}=1}^{m} \min_{j:\mathbf{p}_j\notin\mathbf{P}_{y_i}}  \text{sim}(\mathbf{p}_{\hat{\jmath}},\mathbf{p}_j) \ .
\end{align}

\subsection{Results with Confidence Intervals}
Due to a lack of space we shifted the results with confidence intervals to the Appendix which you can see in tab.~\ref{tab:exp_performance_appendix}.
\begin{table}[t]
    \small
	\centering
	{\def\arraystretch{1.}\tabcolsep=2.pt
	\begin{tabular}{c|l|c|c|c}
	    \hline
		\textbf{a)} &Language Model & Yelp & Movie & Toxicity \\
		\hline
		 \parbox[t]{2mm}{\multirow{4}{*}{\rotatebox[origin=c]{90}{sent.-level}}}&SBERT & $94.92 {\scriptstyle\pm0.01}$ & $84.56{\scriptstyle\pm0.91}$ & $\circ\mathbf{84.40}{\scriptstyle\pm0.03}$ \\
		&SBERT (Proto-Trex) & $93.59{\scriptstyle\pm0.16}$ & $80.05{\scriptstyle\pm0.26}$ & $73.19{\scriptstyle\pm0.71}$ \\
		&CLIP & $93.78 {\scriptstyle\pm 0.00}$ & $75.49{\scriptstyle\pm0.21}$ & $80.82{\scriptstyle\pm0.28}$ \\
		&CLIP (Proto-Trex) & $87.16 {\scriptstyle\pm1.56}$ & $63.52 {\scriptstyle\pm0.66}$ & $67.75 {\scriptstyle\pm2.10}$ \\
		\hline
		 \parbox[t]{2mm}{\multirow{6}{*}{\rotatebox[origin=c]{90}{word-level}}}&BERT & $89.41 {\scriptstyle\pm2.01}$ & $61.45 {\scriptstyle\pm1.16}$ & $76.88 {\scriptstyle\pm1.33}$ \\
		&BERT (Proto-Trex)& $92.10 {\scriptstyle\pm0.08}$ & $75.51 {\scriptstyle\pm0.42}$ & $79.35 {\scriptstyle\pm1.09}$\\
		&GPT-2  & $93.78{\scriptstyle\pm0.41}$ & $\bullet\mathbf{87.05}{\scriptstyle\pm0.31}$ & $81.56{\scriptstyle\pm0.58}$\\
		&GPT-2 (Proto-Trex)& $\bullet\mathbf{95.32}{\scriptstyle\pm0.06}$ & $84.57{\scriptstyle\pm0.31}$ & $84.14{\scriptstyle\pm0.88}$\\
		&DistilBERT & $92.91{\scriptstyle\pm0.07}$ & $79.62{\scriptstyle\pm0.13}$ & $81.57{\scriptstyle\pm0.18}$ \\
		&DistilBERT (Proto-Trex) & $92.71{\scriptstyle\pm0.03}$ & $78.64{\scriptstyle\pm0.14}$ & $83.39{\scriptstyle\pm0.47}$\\
		\hline
		 \parbox[t]{2mm}{\multirow{2}{*}{\rotatebox[origin=c]{90}{inter.}}}&SBERT (iProto-Trex) & $93.81{\scriptstyle\pm0.03}$ & $80.24{\scriptstyle\pm0.31}$ & $73.36{\scriptstyle\pm0.78}$ \\
		&GPT-2 (iProto-Trex) & $\circ\mathbf{95.25}{\scriptstyle\pm0.11}$ & $\circ\mathbf{84.80}{\scriptstyle\pm0.17}$ & $\bullet\mathbf{84.51}{\scriptstyle\pm0.93}$ \\
		\hline
	\end{tabular}
	}
	\newline
    \vspace*{1mm}
    \newline
	{\def\arraystretch{1.}\tabcolsep=2.pt
	\begin{tabular}{l||c|c||c|c}
	    \hline
		\textbf{b)} & \multicolumn{2}{c||}{Word-Selection} & \multicolumn{2}{c}{Similarity} \\
		\hline
		\hline
		Proto-Trex LM & Conv. & Attn. & Cos. & L2 \\
		\hline
		BERT & $\mathbf{92.10}{\scriptstyle\pm0.08}$ & $89.66{\scriptstyle\pm0.92}$ & $\mathbf{92.10}{\scriptstyle\pm0.08}$ & $91.34{\scriptstyle\pm0.28}$\\
		GPT-2 & $\mathbf{95.32}{\scriptstyle\pm0.06}$ & $94.16{\scriptstyle\pm0.07}$ & $\mathbf{95.32}{\scriptstyle\pm0.06}$ & $94.99{\scriptstyle\pm0.09}$\\ 
		DistilBERT & $\mathbf{92.71}{\scriptstyle\pm0.07}$ & $91.09{\scriptstyle\pm0.51}$ &$\mathbf{92.71}{\scriptstyle\pm0.07}$& $92.05{\scriptstyle\pm0.39}$\\ 
		\hline
		SBERT & -- & -- & $\mathbf{93.59}{\scriptstyle\pm0.16}$ & $93.13{\scriptstyle\pm0.34}$ \\ 
		CLIP & --& -- & $\mathbf{87.16}{\scriptstyle\pm1.56}$ & $86.88{\scriptstyle\pm1.47}$\\
		\hline
	\end{tabular}
	}
	\newline
    \vspace*{1mm}
    \newline
	{\def\arraystretch{1.}\tabcolsep=2.pt
	\begin{tabular}{l||c|c||c|c}
	    \hline
		\textbf{c)} & \multicolumn{2}{c||}{Faithfulness} & \multicolumn{2}{c}{Acc.} \\
		\hline
		\hline
		Proto-Trex LM & Comp. & Suff. & before & after \\
		\hline
		SBERT & $0.22$ & --$0.08$ & $80.05$ & $69.66$\\
		iSBERT & $0.21$ & --$0.08$ & $80.24$ & $70.76$\\ 
		\hline
		GPT-2 & $0.12$ & $0.02$ & $84.57$ & $49.91$ \\
		iGPT-2 & $0.13$ & $0.02$ & $84.80$ & $55.74$ \\ 
		\hline
	\end{tabular}
	}
	\newline
    \vspace*{1mm}
    \newline
	\caption{Extension of Tab.~\ref{tab:exp_performance} with respective confidence intervals.}
	\label{tab:exp_performance_appendix}
\end{table}
\subsection{Projection}
We show the impact of projecting the prototypes onto their nearest neighbor in Tab.~\ref{tab:exp_projection_appendix}. The projection is evaluated for sentence-level Proto-Trex networks. One can see the trade-off between interpretability and accuracy introduced by projection.
\begin{table*}[t]
    \small
	\centering
	{\def\arraystretch{1.}\tabcolsep=3.pt
	\begin{tabular}{l|c|c|c}
	    \hline
		Language Model & Yelp & Movie & Toxicity \\
		\hline
		SBERT (Proto-Trex) & $94.13{\scriptstyle\pm0.15}$ & $83.23{\scriptstyle\pm0.05}$ & $83.11{\scriptstyle\pm0.19}$ \\
		SBERT (Proto-Trex) with projection & $93.59{\scriptstyle\pm0.16}$ & $80.05{\scriptstyle\pm0.26}$ & $73.19{\scriptstyle\pm0.71}$ \\
		CLIP (Proto-Trex) & $93.41{\scriptstyle\pm0.08}$ & $73.62{\scriptstyle\pm0.20}$ & $80.74{\scriptstyle\pm0.01}$ \\
		CLIP (Proto-Trex) with projection & $87.16{\scriptstyle\pm1.56}$ & $63.52{\scriptstyle\pm0.66}$ & $67.75{\scriptstyle\pm2.10}$ \\
		\hline
	\end{tabular}
	}
	\caption{Impact of Projection on Proto-Trex networks. The mean balanced accuracy (5 runs) is given in percent with confidence intervals.}
	\label{tab:exp_projection_appendix}
\end{table*}

\subsection{Provided Explanations}
Here we show the explanations that Proto-Trex provides. For each network, we use $10$ prototypes, and for word-level networks, we use a prototype length of $4$ tokens. First we extend the results of Tab.~\ref{tab:exp_query_word} for the word-level case, shown in Tab.~\ref{tab:word-level-query}. Then we show prototype lists for the models on different datasets. In Tab.~\ref{tab:proto_list_gpt2} we show all prototypes for Proto-Trex based on the GPT-2 transformer and in Tab.~\ref{tab:proto_list_sentbert} the prototypes based on the SBERT transformer. Both tables show the prototypes for the Yelp Open and MovieReview dataset. 
The results for the Jigsaw Toxicity dataset can be found in an external document in the codebase. \mbox{{\fontencoding{U}\fontfamily{futs}\selectfont\char 66\relax} CONTENT WARNING:} the content in this document can be disturbing due to highly toxic texts! These results do not represent the authors' opinion and show prototypical explanations provided solely by Proto-Trex.

Tab.~\ref{tab:proto_list_sent_pruned} shows the pruned prototypes of iProto-Trex from Tab.~\ref{tab:proto_list_sentbert}(a). Pruning cuts off the words at the end of a sequence that go beyond two sentences or $15$ tokens in total.

We showcase all experiments of the user interaction in Tab.~\ref{tab:exp_interaction_appendix} which is an extension of Tab.~\ref{tab:exp_interaction}. The extension includes (1) retraining the classification layer for the same number of epochs to provide a fairer comparison with the interaction methods and to exclude changes in accuracy simply due to training for more epochs, (2) pruning the original, (3) removing the original, (4) adding a new prototype without changing the original, (5) replacing the original with the user-chosen alternative ``They offer a bad service.'' without using the interaction loss and (6) fine-tuning the original prototype.

In Fig.~\ref{tab:motivation_query_appendix} we shows the full explanations with importance scores for the exemplary query in the motivation (\cf Fig.~1). It highlights the advantage of using prototype networks and the benefit of interactive learning.

\begin{table*}[t]
    \small
	\centering
	{\def\arraystretch{1.5}\tabcolsep=2.pt
	\begin{tabularx}{\textwidth}{c|c|X}
	    & Importance & Query: I really like \colorbox{coralpink}{the good food and} \colorbox{orange}{kind service here.}\\
		\hline\hline
		\parbox[t]{3mm}{\multirow{4}{*}{\rotatebox[origin=c]{90}{word-level}}} &${\scriptstyle0.43 \cdot 17.13 =} \mathbf{7.37}$ & \textit{P6:} Excellent baby back ribs. Creamed \colorbox{coralpink}{corn was terrific too}.  Came in the afternoon and was easy to get in and have a relaxing meal.  Excellent service too!!\\
		&${\scriptstyle0.56 \cdot 12.14 =} \mathbf{6.80}$ & \textit{P4:} Food \colorbox{orange}{was great...service} was excellent! Will be eating there regularly.\\
		&${\scriptstyle0.39 \cdot 12.89 =} \mathbf{5.03}$& \textit{P10:} A great experience each \colorbox{orange}{time I come in}. The employees are friendly and the food is awesome.\\
	\end{tabularx}
	}
	\caption{Provided Explanations by Proto-Trex networks. 
	Highlighted words mark matching subsequences between query (top-row) and corresponding top three most similar prototypes (word-level). Importance scores (left, bold) are calculated with cosine similarity and classification weight.
	}
	\label{tab:word-level-query}
\end{table*}

\begin{table*}
    \small
	\begin{subtable}[t]{\linewidth}
		\centering
		\begin{tabularx}{\linewidth}{l|X}
			\rowcolor{LightGray}P1 &  This places not authentic and \colorbox{orange}{the pho is not} that good. Very small uncomfortable dining area and the service was horrible.\\
			P2 & Great ambiance with a menu that is \colorbox{orange}{short and sweet}. The food was delicious and I loved the ginger beer! Excellent service. Definitely recommend this place!\\
			\rowcolor{LightGray}P3 & Waited over hour for food.  \colorbox{orange}{Only one person} making sushi.  Left without food, they still charged for the one beer we had.  Bad server + no food = horrible exp. \\
			P4 & Food \colorbox{orange}{was great...service} was excellent! Will be eating there regularly.\\
			\rowcolor{LightGray}P5 & Called it. This place didn't \colorbox{orange}{stand a chance with} terrible management, gross food and slow service.  Too bad for the workers. \\
			P6 & Excellent baby back ribs. Creamed \colorbox{orange}{corn was terrific too}.  Came in the afternoon and was easy to get in and have a relaxing meal.  Excellent service too!! \\
			\rowcolor{LightGray}P7 & Don't waste your money at this \colorbox{orange}{restaurant. Took} my daughter to celebrate her birthday. They didn't even sing Happy Birthday. The food was horrible.\\
			P8 & \colorbox{orange}{Great food, great} service. Chicken tikka was awesome with fresh peppers. Iced tea had a surprising and refreshingly different flavor.\\
			\rowcolor{LightGray}P9 & \colorbox{orange}{Worst customer service}. No one even said hello after going in 3 separate times. I am an avid spender and have spent lots of money there in the past. Will not shop there again! \\
			P10 & A great experience each \colorbox{orange}{time I come in}. The employees are friendly and the food is awesome. \\		
		\end{tabularx}
		\caption{Yelp}
	\end{subtable}
	\begin{subtable}[t]{\linewidth}
		\centering
		\begin{tabularx}{\linewidth}{l|X}
			\rowcolor{LightGray}P1 & There's too much falseness to the second half, and what began as an intriguing look at youth \colorbox{orange}{fizzles into a} dull, ridiculous attempt at heart-tugging.\\
			P2 & Parts of the film feel a bit too much like an infomercial for ram dass's latest book aimed at the boomer demographic. But mostly it's a work that, with humor, warmth, and intelligence, \colorbox{orange}{captures a life} interestingly lived.\\
			\rowcolor{LightGray}P3 & Godard's ode to tackling life's wonderment is a rambling and incoherent manifesto about the vagueness of topical excess... In praise of love remains a ponderous and pretentious \colorbox{orange}{endeavor that's unfocused} and tediously exasperating. \\
			P4 & \colorbox{orange}{It's a lovely} film with lovely performances by buy and accorsi. \\
			\rowcolor{LightGray}P5 & It throws quirky characters, odd situations, and off-kilter dialogue at us, all as if to say, "Look at this! This is an interesting movie!" but the film itself is \colorbox{orange}{ultimately quite unengaging}.\\
			P6 & Like a Tarantino \colorbox{orange}{movie with heart,} alias Betty is richly detailed, deftly executed and utterly absorbing.\\
			\rowcolor{LightGray}P7 & \colorbox{orange}{A bland, obnoxious} 88-minute infomercial for universal studios and its ancillary products.\\
			P8 & The film fearlessly gets under the \colorbox{orange}{skin of the people} involved... this makes it not only a detailed historical document, but an engaging and moving portrait of a subculture. \\
			\rowcolor{LightGray}P9 & A sad and rote exercise in milking a played-out idea -- a straight guy has to dress up in drag -- \colorbox{orange}{that shockingly manages to} be even worse than its title would imply. \\
			P10 & \colorbox{orange}{an inventive, absorbing} movie that's as hard to classify as it is hard to resist. \\		
		\end{tabularx}
		\caption{Movie}
	\end{subtable}
	\caption{Proto-Trex GPT-2 Prototypes. The prototypes are received with GPT-2 for the Yelp Open (a) and MovieReview (b) dataset. The prototypical subsequences provided by GPT-2 are highlighted in color.}
	\label{tab:proto_list_gpt2}
\end{table*}

\begin{table*}
    \small
	\begin{subtable}[t]{\linewidth}
		\centering
		\begin{tabularx}{\linewidth}{l|X}
			\rowcolor{LightGray}P1 & This place is horrible. The staff is rude and totally incompetent. Jose was horrible and is a poor excuse for a customer representative.\\
			P2 & This place is really high quality and the service is amazing and awesome! Jayden was very helpful and was prompt and attentive. Will come back as the quality is so good and the service made it!\\
			\rowcolor{LightGray}P3 & Terrible delivery service. People are mean and don't care about their customers service. I will not ever come back to this place. Also the food is small, not very good and too expensive. \\
			P4 & Incredibly delicious!!! Service was great and food was awesome. Will definitely come back! \\
			\rowcolor{LightGray}P5 & They sat us 30 minutes late for our reservation and didn't get our entree for 1.5 hours after being seated.  The service was terrible. \\
			P6 & I found this place very inviting and welcoming. The food is great so as the servers.. love the food and the service is phenomenal!! Well done everyone..!\\
			\rowcolor{LightGray}P7 & Horrible customer service. Not helpful at all and very rude. Very disappointed and will not go back to this location.\\
			P8 & Oliver Rocks! Great hidden gem in the middle of Mandalay Bay! Great friends great times\\
			\rowcolor{LightGray}P9 & They literally treat you like you are bothering them.  No customer service skills.  Substandard work. \\
			P10 & Fun and friendly atmosphere, fantastic selection. The sushi is so fresh and the flavors are WOW. \\		
		\end{tabularx}
		\caption{Yelp}
	\end{subtable}
	\begin{subtable}[t]{\linewidth}
		\centering
		\begin{tabularx}{\linewidth}{l|X}
			\rowcolor{LightGray}P1 &...in the pile of useless actioners from mtv schmucks who don't know how to tell a story for more than four minutes.\\
			P2 &  the solid filmmaking and convincing characters makes this a high water mark for this genre.\\
			\rowcolor{LightGray}P3 & dull, lifeless, and amateurishly assembled. \\
			P4 & it's a wise and powerful tale of race and culture forcefully told, with superb performances throughout. \\
			\rowcolor{LightGray}P5 & stale, futile scenario. \\
			P6 & a real winner -- smart, funny, subtle, and resonant.\\
			\rowcolor{LightGray}P7 & plodding, poorly written, murky and weakly acted, the picture feels as if everyone making it lost their movie mojo.\\
			P8 & an enthralling, entertaining feature. \\
			\rowcolor{LightGray}P9 & fails in making this character understandable, in getting under her skin, in exploring motivation... well before the end, the film grows as dull as its characters, about whose fate it is hard to care. \\
			P10 & this delicately observed story, deeply felt and masterfully stylized, is a triumph for its maverick director. \\		
		\end{tabularx}
		\caption{Movie}
	\end{subtable}
	\caption{Proto-Trex SBERT Prototypes. The prototypes are received with SBERT for the Yelp Open (a) and MovieReview (b) dataset.}
	\label{tab:proto_list_sentbert}
\end{table*}

\begin{table*}
    \small
	\centering
	\begin{tabularx}{\linewidth}{l|X}
		\rowcolor{LightGray}P1 &This place is horrible. The staff is rude and totally incompetent. Jose was horrible and is\\
		P2 & This place is really high quality and the service is amazing and awesome! Jayden was\\
		\rowcolor{LightGray}P3 & Terrible delivery service. People are mean and don't care about their customers service. I will \\
		P4 & Incredibly delicious!!! Service was great and food was awesome. Will definitely come back! \\
		\rowcolor{LightGray}P5 & They sat us 30 minutes late for our reservation and didn't get our entree for\\
		P6 & I found this place very inviting and welcoming. The food is great so as the servers..\\
		\rowcolor{LightGray}P7 & Horrible customer service. Not helpful at all and very rude. Very disappointed and will not go\\
		P8 & Oliver Rocks! Great hidden gem in the middle of Mandalay Bay! \\
		\rowcolor{LightGray}P9 & They literally treat you like you are bothering them.  No customer service skills.  Substandard work. \\
		P10 & Fun and friendly atmosphere, fantastic selection. The sushi is so fresh and the flavors are \\		
	\end{tabularx}
	\caption{Pruning Prototypes with iProto-Trex. It shows the prototypes from Tab.~\ref{tab:proto_list_sentbert}(a) after pruning has been applied. Pruning reduces the sequences in length while preserving the sentiment. The accuracy remains the same (\cf Tab.~\ref{tab:exp_interaction_appendix}). Prototypes 4 and 9 are not pruned as this would alter the sentiment too much.}
	\label{tab:proto_list_sent_pruned}
\end{table*}

\begin{table*}[t]
    \small
	\begin{tabularx}{\textwidth}{c|c|X}
		Method & Acc. & Prototype \\
		\hline\hline
		no interaction & $93.64$ & Horrible customer service and service does not care about safety features. That's all I'm going to say. Oh they also don't care about their customers\\
		\hline
		retrain & $93.64$ & Horrible customer service and service does not care about safety features. That's all I'm going to say. Oh they also don't care about their customers\\\hline
		prune & $93.64$ & Horrible customer service and service does not care about safety features. That's all I 'm\\\hline
		remove & $93.61$ & -- \\\hline
		add & $93.79$ & They offer a bad service. \\\hline
		replace & $93.78$ & They offer a bad service. \\\hline
		re-initialize & $93.80$ & Terrible delivery service. People are mean and don't care about their customers service. I will not ever come back to this place. Also the food is small, not very good and too expensive. \\\hline
		fine-tune & $93.81$ & Terrible delivery service. People are mean and don't care about their customers service. I will not ever come back to this place. Also the food is small, not very good and too expensive. \\\hline
		soft replace ($0.5$)& $93.81$ & Terrible delivery service. People are mean and don't care about their customers service. I will not ever come back to this place. Also the food is small, not very good and too expensive. \\\hline
		soft replace ($0.9$)& $93.79$ & I really don't recommend this place. The food is not good, service is bad. The entertainment is so cheesy. Not good \\\hline
		soft replace ($1.0$) & $93.79$ & They offer a bad service. \\
	\end{tabularx}
	\caption{User Interaction with iProto-Trex. Each row shows a different interaction method with the balanced accuracy on the test set conducted on Yelp Open dataset. The interaction methods are able to remove the unwanted prototype while incorporating more knowledge and hence more interpretability along with a slightly higher accuracy. This setup illustrates our bidirectional interaction loop.}
	\label{tab:exp_interaction_appendix}
\end{table*}

\begin{table*}[t]
    \small
	\centering
	\begin{tabularx}{\linewidth}{c|c|X}
		&Importance & Query: Staff is available to assist, but limited to the knowledge and understanding of the individual \\
		\hline\hline
		\multirow{6}{*}{\rotatebox[origin=c]{90}{Proto-Trex}}&${\scriptstyle0.58 \cdot 6.95 =} \mathbf{4.01}$ & They literally treat you like you are bothering them.  No customer service skills. Substandard work. \\
		&${\scriptstyle0.38 \cdot 2.61 =} \mathbf{1.00}$ & They sat us 30 minutes late for our reservation and didn't get our entree for 1.5 hours after being seated. The service was terrible. \\
		&${\scriptstyle0.36 \cdot 2.02 =} \mathbf{0.73}$ & This place is horrible. The staff is rude and totally incompetent. Jose was horrible and is a poor excuse for a customer representative. \\
		\hline
		\multirow{5}{*}{\rotatebox[origin=c]{90}{iProto-Trex}}&${\scriptstyle0.50 \cdot 3.25 =} \mathbf{1.63}$ & They offer a bad service. \\
		&${\scriptstyle0.38 \cdot 4.17 =} \mathbf{1.58}$ & They sat us 30 minutes late for our reservation and didn't get our entree for 1.5 hours after being seated. The service was terrible. \\
		&${\scriptstyle0.36 \cdot 3.01 =} \mathbf{1.08}$ & This place is horrible. The staff is rude and totally incompetent. Jose was horrible and is a poor excuse for a customer representative. \\
	\end{tabularx}
	\caption{Top-3 Explanations for Query in the Introduction (Fig.~\ref{fig:motivation}). The first three rows depict explanations with sentence-level Proto-Trex with the SBERT transformer, while the last three show the top three explanations iProto-Trex. On the left side, one can see the importance scores for the classification.}
	\label{tab:motivation_query_appendix}
\end{table*}

\end{document}